\documentclass{article}
\usepackage{graphicx} % Required for inserting images
\usepackage{natbib}
\usepackage{authblk}
\usepackage{float}
\usepackage{graphicx}
\usepackage{hyperref}
\usepackage{xcolor}

\usepackage{siunitx}
\usepackage{subcaption}
\usepackage{multirow}
\title{Self-Supervised Learning of Synapse Types from EM Images}
\author[1]{Aarav \nolinebreak V \nolinebreak Shetty}
\author[1]{Gary \nolinebreak B \nolinebreak Huang}
\affil[1]{Janelia Research Campus, Howard Hughes Medical Institute, USA}
\date{August 2025}

\begin{document}
\maketitle

\begin{abstract}
Separating synapses into different classes based on their appearance in EM images has many applications in biology.  Examples may include assigning a neurotransmitter to a particular class, or separating synapses whose strength can be modulated from those whose strength is fixed.  Traditionally, this has been done in a supervised manner, giving the classification algorithm examples of the different classes.  Here we instead separate synapses into classes based only on the observation that nearby synapses in the same neuron are likely more similar than synapses chosen randomly from different cells.  We apply our methodology to data from {\it Drosophila}.  Our approach has the advantage that the number of synapse types does not need to be known in advance.  It may also provide a principled way to select ground-truth that spans the range of synapse structure.
\end{abstract}

\section{Introduction}
Separating synapses into different classes based on their appearance in EM images has many applications in biology.  Examples include assigning a neurotransmitter to a particular class or separating synapses whose strength can be modulated from those whose strength is fixed.  

Initial machine learning techniques concentrated on detecting synapses, not classifying them.  Even this is not simple, so it was somewhat of a surprise that synapses could not only be identified, but also divided into classes based on their putative neurotransmitter\citep{eckstein2024neurotransmitter}.  This classification is not obvious to humans from the image data, and significant work has gone into trying to understand what features the machine learning is using to make this distinction\citep{adjavon2024quantitative}.

Traditionally, both these tasks have been done in a supervised manner, giving the classification algorithm examples of the different classes: synapses/not synapses, or examples with known neurotransmitters. Supervised learning requires labeled ground-truth, meaning that each training example must have an associated label describing its class. In contrast, self-supervised learning eliminates the need for explicit labels by using the data itself to generate a learning signal. Contrastive methods used in our research, such as the NT-Xent loss function, train models to bring representations of similar inputs closer together while pushing apart dissimilar ones \citep{chen2020simpleframeworkcontrastivelearning}.

Here, we instead separate synapses into classes based only on the observation that nearby synapses in the same neuron are likely more similar than synapses chosen randomly from different cells.  We apply this approach and analyze the resulting learned representations in {\it Drosophila}, a model system for which extensive ground-truth is known.

Self-supervised classification of synapse types has not been previously attempted, but it potentially has several advantages and uses.  The main advantage is that ground-truth is not required.  Even the number of synapse types does not need to be known in advance.  Another plus is that supervised methods seem quite sensitive to imaging conditions and the exact species used.  This often means that ground-truth cannot be transferred even to closely related data - for example, ground-truth for one species of {\it Drosophila}, or even one particular dataset, cannot be used to classify data from another.  

Self-supervised classification may also provide a principled way to select ground-truth that spans the range of synapse structure.  This range, illustrated through techniques such as UMAP, can show what fraction of useful structural differences are covered by the ground-truth.  Selecting examples in portions of this space that are not well covered may be particularly helpful in improving classification among the supervised classifiers.  This could be particularly useful when working with novel species, where ground-truth can be expensive to acquire (perhaps requiring biochemical and genetic methods).
\section{Previous work}
\begin{figure}[htb]
\begin{center}
\noindent
\includegraphics[width=\textwidth]{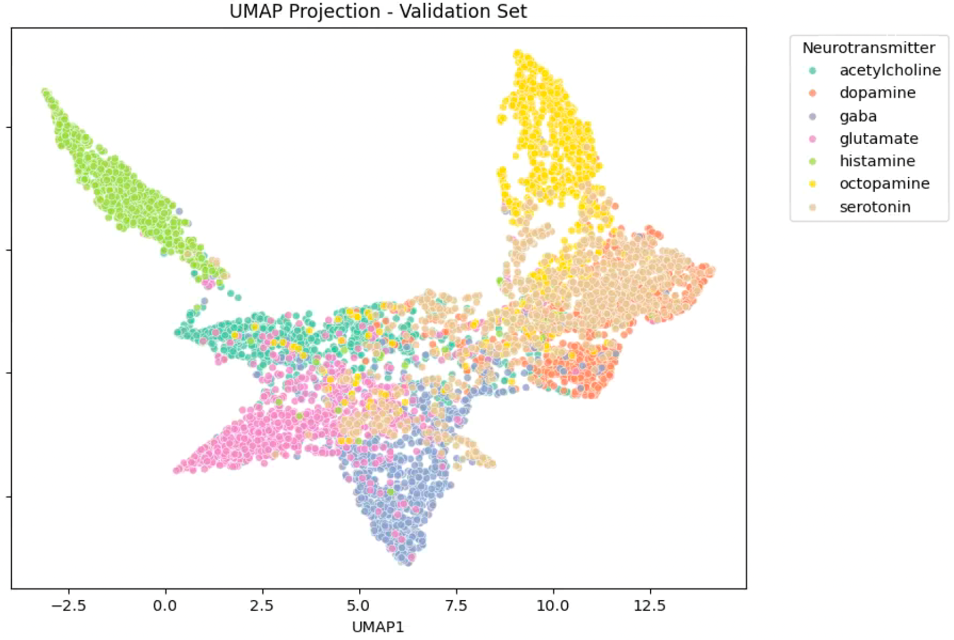}
\caption{Synapses classified by neurotransmitter type.  Two dimensional UMAP reduction of classification output.  Each neurotransmitter is given a distinct color.}
\label{Supervised}
\end{center}
\end{figure}

The current state of the art in synapse classification is shown in Fig. \ref{Supervised}.  This is the result of a supervised classification of synapse types, where the classification algorithm is given ground-truth consisting of synapses of known transmitter types.
This ground-truth is difficult and expensive to obtain, requiring genetic and biochemical evidence for the transmitters of a particular cell type.  In  particular it requires at least a few cells of every transmitter type we hope to separate.  This can be a serious limitation for transmitter types that are used only a few less numerous cells.

\section{Methods}
Dales's principle\citep{dale1935pharmacology}, which is actually a heuristic, specifies that all the synapses of the same neuron use the same neurotransmitter(s).  Furthermore, segmentation divides an EM volume into {\it supervoxels}, which are sets of voxels that segmentation is very sure belong to the same neuron. Therefore, from a classification viewpoint, we would expect all synapses from the same supervoxel to share the same classification.

We rely on having a segmentation that is accurate and conservative, meaning it should contain as few false merges between different neurons as possible. A false merge occurs when parts of two distinct neurons are incorrectly joined into a single segment, leading to mixed neuronal identities and potentially misleading classifications. To achieve accurate segmentation, we use the Flood-Filling Networks method developed by Google, which is designed to generate precise segmentations with minimal false merges in large-scale EM datasets \citep{Januszewski200675}.

Conversely, we would expect that comparing synapses from different supervoxels would result in a very different distribution.  Some will be in the same class (those that share the same transmitter), but others will differ. The question then is whether self-supervised machine learning can learn this distinction. 

In order to examine each synapse, we used a Visual Geometry Group(VGG) model \citep{simonyan2014very}. We used an NT\_Xent loss function for the Self-Supervised model \citep{chen2020simpleframeworkcontrastivelearning}. The function essentially gives a low loss value when two neurotransmitters that are known to be the same are mapped to similar representations in the learned high dimensional space.

The network had an input receptive field size of 80\(^3\)voxels, of monochrome images taken for the purpose of EM connectomics, with an isotropic 8x8x8 nm resolution.  One of the main advantages of this data is that existing neurotransmitter ID had already been run, providing a way to test our results. 
Each subvolume was centered on a synapse in the volume, as determined by a previous step of synapse detection.  In addition, all synapses were labeled with the ID of the supervoxel that contained them.

We trained this network on an EM dataset of the male brain of {\it Drosophila},  the `CNS' dataset\citep{Berg2025.10.09.680999}. 'CNS' has an image volume of 160 teravoxels. The dataset contains 166,696 proofread brain and ventral nerve cord neurons. 

\section{Results}
\begin{figure}[htb]
\begin{center}
\noindent
\includegraphics[width=\textwidth]{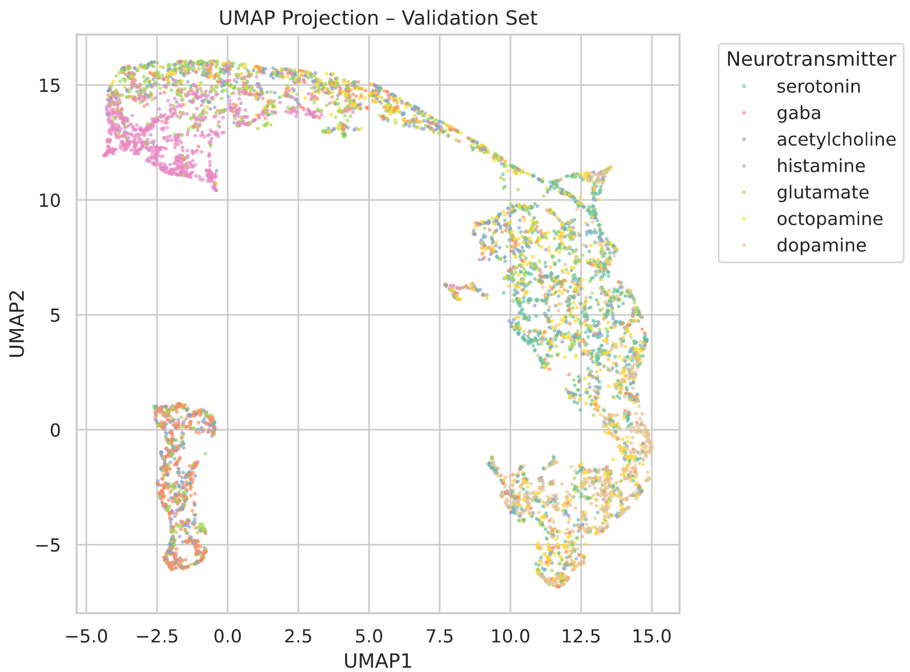}
\caption{Results of self-supervised classification using only Dale's principal and segmentation.}
\label{fig:Self-supervised}
\end{center}
\end{figure}

 We achieve a self-supervised separation as shown in Fig. \ref{fig:Self-supervised}. To visualize the structure of the learned representations, we applied Uniform Manifold Approximation and Projection(UMAP) \citep{mcinnes2020umapuniformmanifoldapproximation}. UMAP is a nonlinear dimensionality reduction technique that reduces high-dimensional data to a low-dimensional space while preserving important structure, making it easier to see patterns. In our case, UMAP was applied to the penultimate representations extracted from trained networks, providing a 2D embedding of the learned synaptic features.

We tested whether this procedure could separate neurotransmitters as intended. To do so, we colored each example synapse in the resulting UMAP according to its neurotransmitter ID. These IDs were determined experimentally from ground-truth data \citep{drosophila_neurotransmitters}. This allowed us to directly assess how well the procedure captured biologically meaningful distinctions.  

Neurotransmitter ID did not appear to be a significant feature, at least according to the UMAP of the two most significant feature axes.  This is clear from comparison of the same UMAP plot generated with explicit ground-truth  (Fig. \ref{Supervised}.) for transmitter ID, where the synapses are divided into groups of different color.

\section{Conclusions and future work}
This work shows that synapses, at least in {\it Drosophila} EM images, can be separated by a combination of self-supervised learning and Dale's principle.  We were hoping that one of the main features leading to this separation would be neurotransmitter ID, but this appears not to be the case.

However, even if this technique cannot reliably determine neurotransmitter IDs, there are possible applications of this work.

First, it should be possible to use the results of our work to initialize the search for neurotransmitter prediction.  Starting from a procedure that can already tell synapses apart should reduce the work needed to train the particular classifier needed.

Second, this work could be used to help automate proofreading. The initial segmentation often contains {\it false merges} where fragments of two neurons have been incorrectly joined.  from Dale's principal, if we find very different synapse types in what is supposed to be single neuron, this is strong evidence that a false merge has taken place, and looking at the boundary between the synapse types can reveal roughly where this false merge happened.

Finally, self-supervised classification may also provide a principled way to select ground-truth that spans the range of synapse structure.  This range, illustrated through techniques such as UMAP, can show what fraction of useful structural differences are covered by the ground-truth.  Selecting examples in portions of this space that are not well covered may be particularly helpful in improving classification among the supervised classifiers.  This could be particularly useful when working with novel species, where ground-truth can be expensive to acquire (perhaps requiring biochemical and genetic methods).

An obvious additional question is what features of the synapses are being recognized in the distinct clumps in the lower left, upper left, and middle right of the UMAP diagram.

\bibliography{auto}
\end{document}